\documentclass[final]{cvpr}

\usepackage{times}
\usepackage{epsfig}
\usepackage{enumitem}
\usepackage[utf8]{inputenc} % allow utf-8 input
\usepackage[T1]{fontenc}    % use 8-bit T1 fonts
\usepackage{url}            % simple URL typesetting
\usepackage{booktabs}       % professional-quality tables
\usepackage{amsfonts}       % blackboard math symbols
\usepackage{nicefrac}       % compact symbols for 1/2, etc.
\usepackage{microtype}      % microtypography
\usepackage{graphicx}
\usepackage{subfigure}
\usepackage{amsmath,amssymb}
\usepackage{mathrsfs}
\usepackage{multirow}
\usepackage{float}
\usepackage{setspace}
\usepackage{mathrsfs}
\usepackage{caption}
\usepackage[pagebackref=true,breaklinks=true,colorlinks,bookmarks=false]{hyperref}

\def\cvprPaperID{3293} % *** Enter the CVPR Paper ID here
\def\confYear{CVPR 2021}

\begin{document}

%%%%%%%%% TITLE
\title{\ RSG: A Simple but Effective Module for Learning Imbalanced Datasets}

\author{Jianfeng Wang\textsuperscript{1}, \ \ 
Thomas Lukasiewicz\textsuperscript{1}, \ \ Xiaolin Hu\textsuperscript{2}, \ \  Jianfei Cai\textsuperscript{3}, \ \  Zhenghua Xu\textsuperscript{4}\thanks{Corresponding author.} \\
University of Oxford\textsuperscript{1} \ \ Tsinghua University\textsuperscript{2} \ \  Monash University\textsuperscript{3} \ \ Hebei University of Technology\textsuperscript{4} \\
{\tt\small \{jianfeng.wang,\,thomas.lukasiewicz\}@cs.ox.ac.uk, xlhu@tsinghua.edu.cn, jianfei.cai@monash.edu} \\
{\tt\small zhenghua.xu@hebut.edu.cn}
}

\maketitle
\pagestyle{empty}  % no page number for the second and the later pages
\thispagestyle{empty} % no page number for the first page

%%%%%%%%% ABSTRACT
\begin{abstract}
   Imbalanced datasets widely exist in practice and are a great challenge for training deep neural models with a good  generalization on infrequent classes. In this work, we propose a new %{\bf R}are-class {\bf S}ample {\bf G}enerator 
   rare-class sample generator (RSG) to solve this problem. 
   RSG aims to generate some new samples for rare classes during training, and it has in particular the following advantages: (1)~it is convenient to use and highly versatile, because it can be easily integrated into any kind of convolutional neural network, and it works well when combined with different loss functions, and (2)~it is only used during the training phase, and therefore, no additional burden is imposed on deep neural networks during the testing phase. 
   %Furthermore, we also propose a new loss function to help with optimizing RSG, namely, maximized vector loss. 
   In extensive experimental evaluations, we verify the effectiveness of RSG. Furthermore, by leveraging RSG, we obtain competitive results on Imbalanced CIFAR and new state-of-the-art results on Places-LT, ImageNet-LT, and iNaturalist 2018. The source code is available at \href{https://github.com/Jianf-Wang/RSG}{https://github.com/Jianf-Wang/RSG}.
\end{abstract}

%%%%%%%%% BODY TEXT
\vspace{-0.4cm}
\section{Introduction}
Computer vision research has made great progress in the past few years, driven by the development of deep convolutional neural networks (CNNs)  \cite{krizhevsky2012imagenet, szegedy2015going, he2016deep, huang2017densely, xie2017aggregated, jf2021grcnn, wang2017gated, Chen_2020_CVPR, hu2018squeeze, tan2019efficientnet} as well as large-scale datasets of high quality \cite{deng2009imagenet, lin2014microsoft}. However, these large-scale datasets are usually well-designed, and the number of instances in each class is balanced artificially, which is inconsistent with the data distributions in real-world scenaries. It is common that the images of some categories are difficult to be collected, resulting in a dataset with an imbalanced data distribution. In general, imbalanced datasets can be classified into two categories in terms of data distributions: long-tailed imbalanced distributions \cite{cui2019class} and step imbalanced distributions \cite{buda2018systematic}, 
which will both be the focus of this work. 

\begin{figure}
\centering
\includegraphics[width=\linewidth]{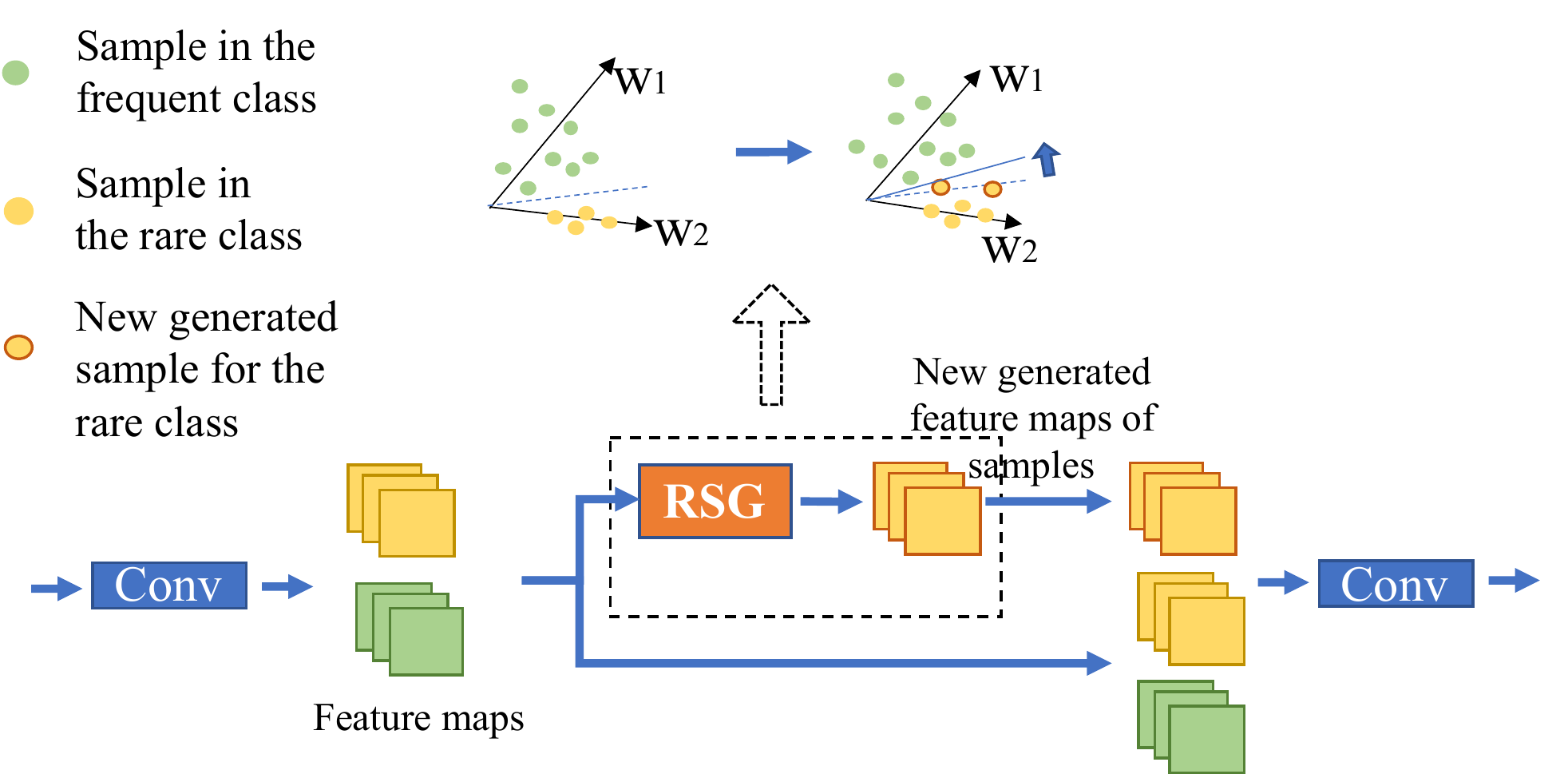}
\caption{RSG in a simple CNN. The part in the dotted
box is only used during training. RSG learns to generate new rare-class samples, which are used to reshape the decision boundary and enlarge the feature space of rare classes.
}
\vspace{-2ex}
\label{fig:cnn_RSG}
\vspace{-1ex}
\end{figure}

Generating new samples for rare classes during training is a good solution \cite{schwartz2018delta, wang2018low, yin2019feature}, which is regarded as a data augmentation method. However, these methods have different drawbacks, which limit their performance. Firstly, some frameworks \cite{schwartz2018delta,yin2019feature} were not trained in an end-to-end manner, so that the gradients cannot be backpropagated from the top to the bottom of CNNs. But it is well known that deep models can usually benefit from  end-to-end training. Secondly, some methods \cite{schwartz2018delta, yin2019feature} utilized variation information, such as different poses or lighting, among samples from the same frequent class to generate new rare-class samples.  However, these methods did not introduce any mechanism to ensure that the variation information obtained from frequent classes is class-irrelevant.  As a result, if the variation information (which still contains the class-relevant information) is directly combined with real rare-class samples to generate new rare-class ones for training the classifier and reshaping decision boundaries, the performance will be hurt due to the aliasing of different class-relevant information. 
Finally, Wang \emph{et al.} \cite{wang2018low} use noise vectors to encode the variation information mentioned above. But using such noise vectors for generation can possibly generate unstable or low-quality samples, since noise vectors are too random to reflect the true variations among real images.\footnote{Note that \cite{yin2019feature} has proposed to avoid sampling random vectors due to their randomness, and  \cite{schwartz2018delta} also has conducted experiments and verified that using random vectors to generate new samples for training classifiers can degrade the performance.}

To alleviate the above drawbacks, in this paper, we propose a simple but efficient fully parameterized generator, called %\bf R}are-class {\bf S}ample {\bf G}enerator 
rare-class sample generator (RSG), which can be trained end-to-end with any backbone. RSG directly uses the variation information, which usually reflects different poses or lighting, among the real samples from the same frequent class to generate new samples rather than using random vectors to encode such information, and therefore, RSG can generate more reasonable and stable samples. 
Besides, RSG introduces a new module that is designed to further filter out the frequent-class-relevant information that possibly exists in the variation information, solving the aliasing problem mentioned above. 

Figure~\ref{fig:cnn_RSG} shows how it is integrated into a simple CNN for imbalanced datasets. RSG only requires the feature maps of samples from any specific layer, and it generates some new samples during training to impact on rare classes in order to adjust their decision boundaries and enlarging their feature space. In the testing phase, RSG is removed, so that no additional computational burden is imposed on the network. Note that we only show a simple CNN in Fig.~\ref{fig:cnn_RSG}, but RSG can be used in any network architecture, such as ResNet \cite{he2016deep}, DenseNet \cite{huang2017densely}, ResNeXt \cite{xie2017aggregated}, and Inception \cite{szegedy2015going}.

\begin{figure*}[t]
 \centering
 \includegraphics[width=0.75\linewidth]{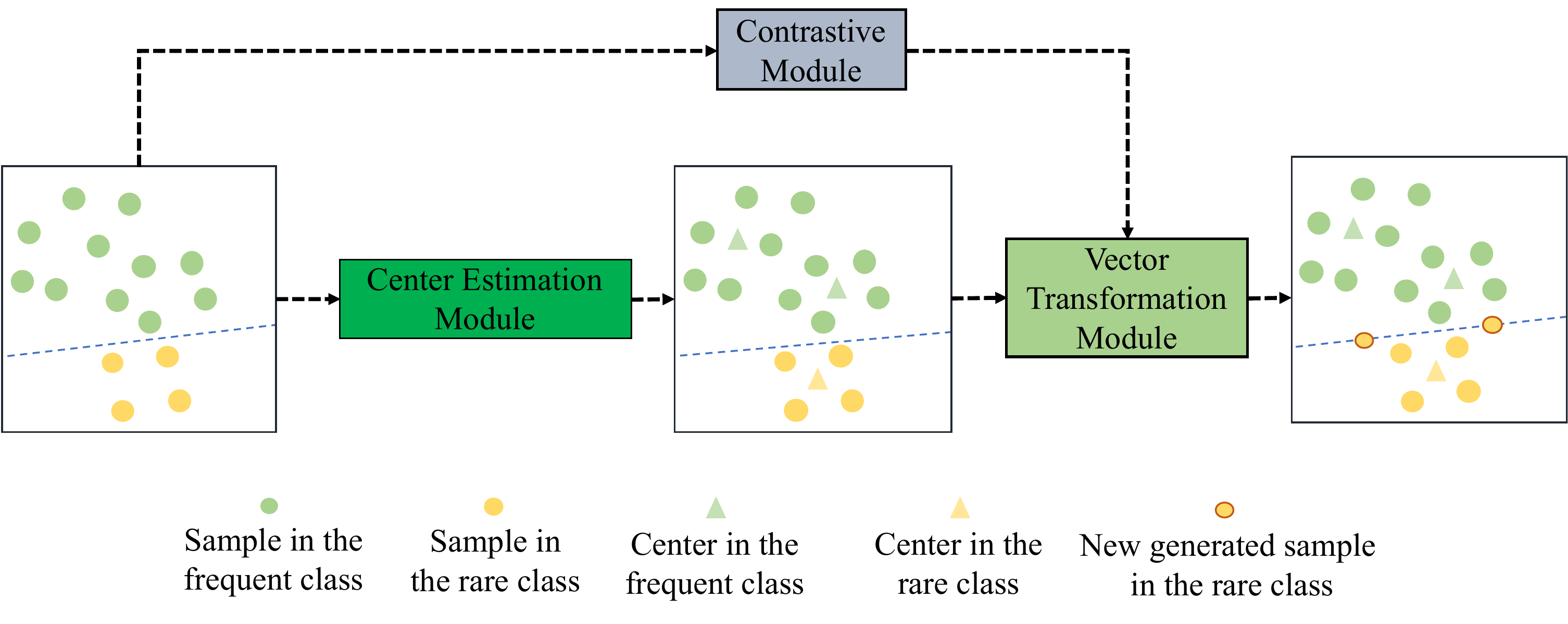}
  \vspace{-0.3cm}
 \caption{A diagram of RSG with samples' feature maps as input. The blue dashed line denotes a decision boundary.}
 \label{fig:RSG}
 \vspace{-1.2ex}
\end{figure*}

\section{Related Work}
\label{related_work}

Recent existing solutions for dealing with imbalanced datasets can be largely classified into approaches based on resampling and reweighting, new loss functions, meta-learning, utilizing unlabeled data, and sample generation. 

Resampling techniques include oversampling the minority classes \cite{shen2016relay, buda2018systematic, byrd2019effect, zhou2020BBN, kang2019decoupling} and undersampling the majority classes \cite{buda2018systematic, japkowicz2002class, he2009learning}, which aims to balance the data distribution. Reweighting methods \cite{huang2016learning,huang2019deep,wang2017learning, cui2019class, li2019gradient} also try to balance the data distribution by assigning different weights to frequent-class and rare-class samples. Some approaches \cite{zhang2017range, cao2019learning} designed new loss functions by directly adding constraints to affect the decision boundaries for frequent and rare classes. Some meta-learning-based methods \cite{liu2019large, snell2017prototypical, shu2019meta} were also proposed to solve the data imbalance problem. Very recently, Yang and Xu \cite{yang2020rethinking} analyzed the value of imbalanced labels, and utilized unlabeled data to boost class-imbalanced learning via semi-supervised and self-supervised strategies. 

Previous sample generation methods are more relevant to this work than other approaches. A hallucinator \cite{wang2018low} was designed to generate new samples for rare classes. It uses real instances from rare classes and noise vectors to produce new hallucinated instances for rare classes. 
A $\Delta$-encoder framework \cite{schwartz2018delta} was proposed  for generating new samples. It is first trained to reconstruct the pre-computed feature vector of input images from frequent classes. Thereafter, it is used to generate new samples by combining the real rare-class samples, and the newly generated ones are further used to train the classifier. 
A feature transfer learning (FTL) framework \cite{yin2019feature} was recently proposed, which consists of an auto-encoder, a feature filter, and fully-connected (FC) layers. The auto-encoder is initially pre-trained on a large-scale dataset for several epochs to converge to learn the latent representations. Then, principal component analysis (PCA) is leveraged to transfer the intra-class variance from frequent classes to rare classes by generating some new rare-class samples. A two-stage alternating training strategy was also proposed to jointly optimize the encoder, the feature filter, and FC layers.

\section{Rare-Class Sample Generator (RSG)} % RSG and MV Loss}
\label{model}
The rare-class sample generator (RSG) is composed of a center estimation module, a contrastive module, and a vector transformation module (see Fig.~\ref{fig:RSG}). 
To~optimize the parameters of RSG, two loss functions are used, namely, center estimation with sample contrastive (CESC) loss
and maximized vector (MV) loss.

RSG assumes that samples from a class follow a uni-modal distribution or a multi-modal distribution \cite{snell2017prototypical, yin2019feature}, and thus there can be a center or a set of centers in each category to fit the distribution. In this paper, we define the notion of \emph{feature displacement}, 
which  indicates the displacement of a sample to its corresponding center in a class,  caused by the same object with different conditions (e.g., angles, poses, or light conditions) in input images. Therefore, under ideal circumstances,  feature displacement should not contain class-relevant information.

Given a mini-batch of samples consisting of both frequent-class and rare-class instances, RSG takes their feature maps as input and forwards them to these modules. 
The center estimation module aims to estimate a set of centers in each class, which is used as ``anchors'' for obtaining the feature displacement of each sample. 
%In this process, those input samples are assigned to the closest centers in their corresponding class to update the centers. 
The contrastive module is used to ensure that the feature displacement does not contain any frequent-class-relevant information during the sample generation process. 
%It receives a pair of samples and performs a binary classification in order to judge whether the two input samples belong to the same category. 
The vector transformation module calculates the {feature displacement} of each frequent-class sample based on the estimated centers and uses it for generating new samples for rare classes.  Intuitively, generating some new samples with such {feature displacement} that comes from abundant classes for rare classes may alleviate the problem caused by imbalanced datasets, as rare classes usually lack input variations.

\medskip 
\noindent\textbf{The center estimation module}
is formulated as:
\begin{equation}
\label{eq:eq0}
\gamma^l =f(A^l ave(x^l) + b^l),
\end{equation}
where $x^l\,{\in}\, R^{D\times W\times H}$ is the feature map of an input sample, and we assume that the channel dimension, width, and height are $D$, $W$, and $H$, respectively.
$l$ is the class label of the sample, $ave(\cdot)$ denotes global average pooling across width and height, $A^l$ and $b^l$ are the parameters of this module  performing a linear transformation on the input,  and $f$ is  the softmax function that outputs a probability distribution ($\gamma^l$) for assigning the sample to the closest center in its corresponding class. 

The center estimation module is designed to estimate a set of centers instead of only one center for each class, since the intra-class data distribution is unknown. If the intra-class data distribution is a multi-modal distribution, using a set of centers is better than using a single center. On the contrary, if it is a uni-modal distribution, those centers can be very close or overlapping, which is similar to using a single center.

\smallskip 
\noindent\textbf{The contrastive module}
%The contrastive module
is formulated as:
\begin{equation}
\label{eq:eq0}
\gamma^* =f(A^*ave(h(cat[x_1, x_2])) + b^*),
\end{equation}
where $x_1{\in}\, R^{D\times W\times H}$ and $x_2{\in}\, R^{D\times W\times H}$ are the feature maps of any two input samples from a given mini-batch, and $cat(\cdot)$ denotes the concatenation operation, which performs along the channel dimension. $h(\cdot)$ is implemented by stacking two $3\times3$ convolutional layers with 256 channels interleaved with a ReLU activation layer throughout the paper. $A^*$ and $b^*$ are the parameters of the linear layer, resulting in a probability distribution $\gamma^*$ to show whether two samples come from the same class.

\medskip 
\noindent\textbf{The vector transformation module} 
is responsible for
generating new rare-class samples through combining the {feature displacement} from real frequent-class samples with real rare-class samples. As Fig.~\ref{fig:cnn_RSG} shows, an imbalanced dataset causes a bias in the decision boundary, resulting in a smaller feature space for rare classes than for frequent classes. 
%In  contrast, in the ideal case, well-balanced datasets lead to an unbiased decision boundary. 
Thus, we propose to use the vector transformation module to generate new samples for rare classes to enlarge the feature space and ``push away'' the decision boundaries.

To generate new samples, we first need to obtain the {feature displacement} from frequent classes, which is implemented by using the frequent-class samples and their corresponding centers estimated by the center estimation module: 
\begin{equation}
 \label{eq:eq4}
  x_{\text{fd-freq}} = x^l_{\text{freq}} -up(C_\mathcal{K}^l),
\end{equation}
where $x^l_{\text{freq}}{\in}\, R^{D\times W\times H}$ denotes a sample in a frequent class~$l$. We use $C_i^l\,{\in}\, R^{D}$ to denote the $i$-th center in class $l$ with dimension $D$, and {\small $\mathcal{K}$} is the index of the closest center to $x^l_{\text{freq}}$, i.e., {\small $\mathcal{K} = \text{arg\,max}\,\, f(A^lave(x_{\text{freq}}^l)+b^l)$.} $up(\cdot)$ denotes the upsampling operation implemented by repeating the values of $C_i^l$ along the width and height, forming feature maps of a center in the same size as the $x^l_{\text{freq}}$. 
After we subtract the corresponding center feature maps from $x^l_{\text{freq}}$, most of the class-relevant information is removed from $x^l_{\text{freq}}$;
thus, we use $x_{\text{fd-freq}}$ to represent the {feature displacement} of the frequent-class sample. 

Then, the second step is to generate new samples for rare classes by using $x_{\text{fd-freq}}$ and the real rare-class samples. Intuitively, $x_{\text{fd-freq}}$ can be added to the centers of rare classes, but we directly add $x_{\text{fd-freq}}$ to the real rare-class samples for two reasons: Firstly, the length of some $x_{\text{fd-freq}}$ may be smaller than the original variance of the feature space in rare classes. If we add $x_{\text{fd-freq}}$ to the centers, the new samples may have no impact on decision boundaries. Secondly, due to the limited sample size of rare classes, most rare-class samples can directly determine the decision boundaries, and adding $x_{\text{fd-freq}}$ to rare-class samples has a more straightforward impact on the decision boundaries.

So, the generation process of new rare-class samples is: 
\begin{equation}
 \label{eq:eq5}
  x_{\text{new}}^{\scriptscriptstyle l'} = \mathcal{T}(x_{\text{fd-freq}}) + x^{\scriptscriptstyle l'}_{\text{rare}},
\end{equation}
where $x^{{\scriptscriptstyle l'}}_{\text{rare}}{\in}\, R^{D\times W\times H}$ denotes a sample in a rare class $l'$,  $x_{\text{new}}^{{\scriptscriptstyle l'}}$ is a newly generated sample in that class, and $\mathcal{T}$ is a linear transformation defined as $\mathcal{T}(z) = conv(z)$, where $conv$ denotes a single convolutional layer containing a set of convolutional filters with the kernel size 3, the stride 1, and the padding size 1, whose number is the same as the number of channels of input feature maps.

\begin{table*}[t]
\centering\resizebox{0.85\textwidth}{!}{
      \begin{tabular}{@{}l@{}}
    \hline
      \textbf{Algorithm 1:} Training Procedure of RSG\\
    \hline
    \textbf{Input: }\\
    Batch size: s; feature maps of training data: $\{x^{(i)}\}_{i=1}^{s}$; epoch threshold: T$_{\text{th}}$; centers: $C$;  training epochs: T;\\
     transfer strength: $\beta\in(0, 1]$; center estimation module: $CE_{\theta}$; contrastive module: $CM_{\theta}$;\\
    vector transformation module: $VT_{\theta}$; weights of the backbone network: $\widetilde{\theta}$; frequent-class ratio: $\alpha\in(0, 1]$.
    \vspace*{1ex}\\
    \textbf{Training:}\\
    \textbf{for} j \textbf{in} range(0, T): \\
    \quad  Compute  $L_{\text{CESC}}$ with $\{x^{(i)}\}_{i=1}^{s}$. Compute gradient $\nabla_{\text{CESC}}$. \\
    \quad Update: $\nabla_{\text{CESC}} \rightarrow CE_{\theta}$, $\nabla_{\text{CESC}} \rightarrow C$.\\
    \quad \textbf{if} j \textless T$_{\text{th}}$: \\
    \quad\quad  Compute  $L_{\text{cls}}$ with $\{x^{(i)}\}_{i=1}^{s}$. Compute gradient $\nabla_{\text{cls}}$. \\     
    \quad\quad Update: $\nabla_{\text{cls}} \rightarrow \widetilde{\theta}$, $\nabla_{\text{CESC}} \rightarrow CM_{\theta}$. \\
    \quad \textbf{else}: \\
    \quad\quad  Generate new samples with $\alpha$ and $\beta$: $\{x_{\text{new}}^{(i)}\}_{i=1}^{ s_{\text{new}}}$. Concat: $\{x_{\text{aug}}^{(i)}\}_{i=1}^{s + s_{\text{new}}}$ = [$\{x^{(i)}\}_{i=1}^{s}$, $\{x_{\text{new}}^{(i)}\}_{i=1}^{ s_{\text{new}}}$]. \\
    \quad\quad  Compute $L_{\text{MV}}$ with $C$, $\{x_{\text{new}}^{(i)}\}_{i=1}^{ s_{\text{new}}}$, and $\{x^{(i)}\}_{i=1}^{s}$. Compute gradient $\nabla_{\text{MV}}$. \\
    \quad\quad Compute $L_{\text{cls}}$ with $\{x_{\text{aug}}^{(i)}\}_{i=1}^{s+ s_{\text{new}}}$. Compute gradient $\nabla_{\text{cls}}$. \\
    \quad\quad Update: $\nabla_{\text{MV}} + \nabla_{\text{cls}} \rightarrow VT_{\theta}$, $\nabla_{\text{cls}} \rightarrow \widetilde{\theta}$. \\
    \quad\textbf{end if} \\
    \textbf{end for} \\
    \hline
  \end{tabular}}
  \vspace{-0.4ex}
       \label{tab:training_proc}
\end{table*}

\begin{figure}[t]
\centering
\includegraphics[width=1.\linewidth]{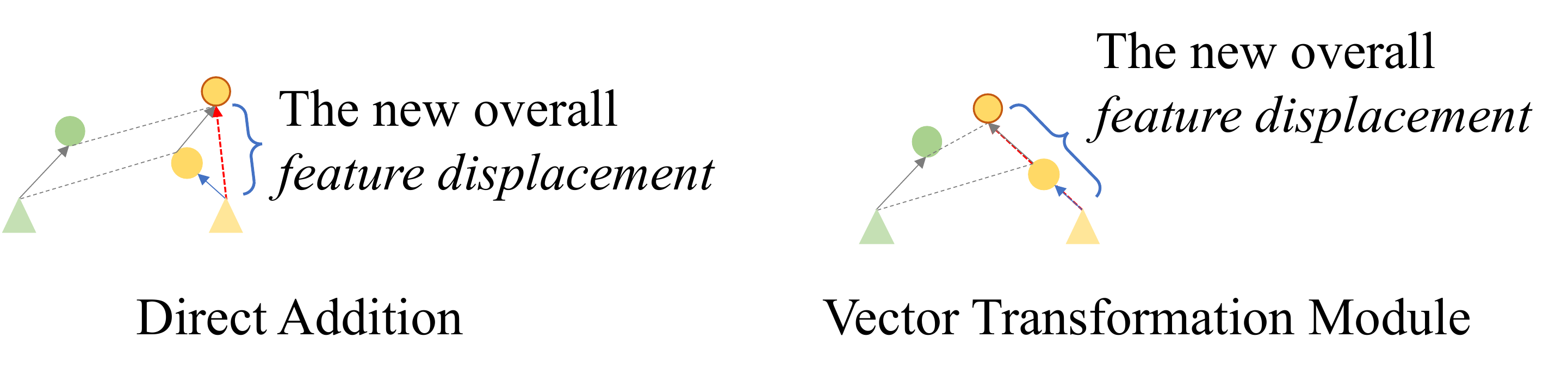}
\caption{The objective and principle of the vector transformation module and MV loss. The triangles and circles in the figure have the same meaning as those in Fig.~\ref{fig:RSG}.}
\label{fig:MV_loss}
\end{figure}

\medskip 
\noindent\textbf{The center estimation with sample contrastive loss
%Center estimation with sample contrastive loss,  
($L_{\text{CESC}}$)} aims to update centers of each class and to optimize the contrastive module as well as the center estimation module. Therefore, it is composed of two classical loss terms, which can be written as:
\begin{equation}
\small
\begin{aligned}
L_{\text{CESC}} = \left \langle \sum_{i=0}^{K-1}\gamma_i^l\sum_{d,j,k}||x^l_{(d,j,k)} - up(C_i^l)_{(d,j,k)}||^2  \right \rangle_s\\ 
- \left \langle (ylog\gamma^* + (1-y)log(1-\gamma^*)) \right \rangle_{\frac{s}{2}},
\end{aligned}
\end{equation}
where $d$, $j$, and $k$ denote the indices of the feature maps along
the channel, width, and height. $\gamma^l_i$ is the probability of the sample belonging to the $i$-th center obtained from Eq.~\eqref{eq:eq0}, $K$ is the number of centers in each class, $s$ is the batch size. 
Considering a mini-batch with batch size $s$, $\frac{s}{2}$ sample pairs are formed by randomly picking samples from the mini-batch during training for the contrastive module. We denote  by $y\in\{0, 1\}$  the ground-truth showing whether the samples in each input pair come from the same class. $\langle \cdot \rangle_s$ and $\langle \cdot \rangle_{\frac{s}{2}}$ denote that the first term and the second term of $L_{\text{CESC}}$ are calculated over $s$ instances and $\frac{s}{2}$ pairs on average,  respectively.

\medskip 
\noindent\textbf{The maximized vector loss
($L_{\text{MV}}$)} optimizes the parameter of the vector transformation module (namely, $\mathcal{T}$) and ensures that newly generated samples can enlarge the feature space of rare classes, where the basic idea is to maximize the {feature displacement} of newly generated samples relative to their centers (i.e., the new overall {feature displacement} in Fig.~\ref{fig:MV_loss}). Here, we treat the {feature displacement} of a sample as a vector starting from a center to the sample (see  Fig.~\ref{fig:MV_loss}). To generate a new rare-class sample, one can directly add $x_{\text{fd-freq}}$ to a rare-class sample (Fig.~\ref{fig:MV_loss}, left), but the direction of $x_{\text{fd-freq}}$ is usually uncertain and the new overall {feature displacement} typically does not always have the largest length, because of the triangle inequality. Thus, we design the MV loss to make the transformed vector co-linear with the {feature displacement} of the rare-class sample in the same direction, and leave the length of the transformed vector unchanged (Fig.~\ref{fig:MV_loss}, right), to maximally impact the decision boundary. 
For example, if the direct addition is used, the newly generated samples may not impact the decision boundary due to the limited overall length. But leveraging the vector transformation module and MV loss ensures that the newly generated samples are widely distributed in the feature space of rare classes, because of the larger displacement relative to the centers, and it improves the probability that newly generated samples can appear around decision boundaries in each batch during training. 

Moreover, as for a given frequent-class sample, although the frequent-class-relevant information has been largely removed when the {feature displacement} of a sample is calculated via Eq.~\eqref{eq:eq4}, we still use the contrastive module to ensure that the {feature displacement} does not contain frequent-class-relevant information in order to further alleviate the possible class-relevant information aliasing problem when new rare-class samples are generated. 
W.l.o.g., $\gamma^*$ is the probability that the two input samples of the contrastive module do not belong to the same category, and the MV loss~is:
\begin{equation}
\label{eq:eq6}
\begin{aligned}
&L_{\text{MV}} = \left \langle \sum_{j,k}(|\frac{\mathcal{T}(x_{\text{fd-freq}})^{(j,k)} \cdot x^{(j,k)}_{\text{fd-rare}}}{||\mathcal{T}(x_{\text{fd-freq}})^{(j,k)}||_2||x^{(j,k)}_{\text{fd-rare}}||_2} - 1|) \right \rangle_{s_{\text{new}}}  \\
&+ \left \langle \sum_{j,k}(|\ ||\mathcal{T}(x_{\text{fd-freq}})^{(j,k)}||_2-||x_{\text{fd-freq}}^{(j,k)}||_2|)\right \rangle_{s_{\text{new}}} \\
& - \left \langle log\gamma^* \right \rangle_{s_{\text{new}}}\,,
\end{aligned}
\end{equation}
where $j$ and $k$ denote the indices of the feature maps along the width and height, $|\cdot|$ takes the absolute value, and $x_{\text{fd-rare}}$ represents the {feature displacement} obtained from a sample and its closest center in a rare class via Eq.~\eqref{eq:eq4}. 
The two input samples of the contrastive module are $\mathcal{T}(x_{\text{fd-freq}})$ and $x^l_{\text{freq}}$, respectively. 
The first term of $L_{\text{MV}}$ is essentially to minimize the cosine angle of $\mathcal{T}(x_{\text{fd-freq}})$ and $x_{\text{fd-rare}}$ in order to make them co-linear in the same direction, the second term is to keep the length of  $\mathcal{T}(x_{\text{fd-freq}})$ unchanged compared with $x_{\text{fd-freq}}$, and the third term makes $\mathcal{T}(x_{\text{fd-freq}})$ and $x^l_{\text{freq}}$ not belong to the same category, ensuring that $\mathcal{T}(x_{\text{fd-freq}})$ will not have any frequent-class-relevant information. 
Given a mini-batch of samples, $\langle \cdot \rangle_{s_{\text{new}}}$ denotes that $L_{\text{MV}}$ is calculated over newly generated samples on averages, where $s_{\text{new}}$ is the number of newly generated samples.

Note that minimizing
$L_{\text{MV}}$ may encourage to
generate some new samples with very large overall {feature displacement}, which can hurt the performance on frequent classes. Thus, the vector transformation module also receives the gradients from the classification loss function $L_{\text{cls}}$, reaching a trade-off between $L_{\text{cls}}$ and the second term of $L_{\text{MV}}$, %Eq.~\eqref{eq:eq6}, 
to generate more reasonable new samples for rare classes.

\medskip 
\noindent\textbf{The training procedure and overall loss function} of RSG
 are summarized in \textbf{Algorithm 1} %that is shown in the appendix in the supplementary material, 
and given as follows, respectively: 
%and the overall loss function of the whole pipeline is: 
%as follows:
\begin{equation}
\label{eq:overall}
L_{total} = L_{\text{cls}} + \lambda_1 L_{\text{CESC}} + \lambda_2 L_{\text{MV}},
  %\vspace{0.2cm}
\end{equation}
where $L_{\text{cls}}$ denotes any classification loss, such as softmax with cross-entropy loss, focal loss \cite{lin2017focal}, AM-Softmax \cite{wang2018additive, wang2018cosface}, and LDAM \cite{cao2019learning}, and $\lambda_1$ and $\lambda_2$ denote coefficients. 
The epoch threshold $T_{\text{th}}$ is set to the index of epoch in which the learning rate is decayed to 0.001 in this paper.

\medskip 
\noindent\textbf{The workflow} of RSG is as follows (see Fig.~\ref{fig:RSG}). Before the epoch threshold $T_{\text{th}}$, given a mini-batch of samples, RSG splits them into two parts according to a manually set constant frequent-class ratio 
%$\alpha$ , which is defined as 
%\begin{equation}
%\small
$\alpha={n_{\text{freq}}}/{n_{\text{cls}}}$,
%\end{equation}
where $n_{\text{freq}}$ and $n_{\text{cls}}$ denote the number of frequent classes and the total number of classes, respectively.
For example, for a training set of 10 classes and $\alpha=0.3$, the three classes with the largest number of samples are frequent classes, and the other classes are rare classes. (Note that for simplicity, only a frequent-class and a rare-class are plotted in Fig.~\ref{fig:RSG}.)
Then, the data are forwarded to the center estimation module to update centers for each class and optimize the parameters of the center estimation module. In addition, those data are also forwarded to the contrastive module to optimize its parameters. 
After the epoch threshold $T_{\text{th}}$, RSG starts to generate new samples and the parameters of the contrastive module are not further updated. The {feature displacement} of each sample in frequent classes is calculated by the vector transformation module, which is then transformed with $\mathcal{T}$ and randomly added to the data in rare classes with a manually set parameter transfer strength $\beta$, resulting in newly generated samples. The contrastive module propagates gradients to the $\mathcal{T}$ in the vector transformation module to optimize $\mathcal{T}$ and filter out frequent-class-relevant information. In general, the number of samples in frequent classes is not smaller than that in rare classes in a given mini-batch. We define the transfer strength $\beta$ as the number of samples in frequent classes involved in calculating the {feature displacement} and generating new samples for rare classes. Specifically, the number of newly generated samples is 
%\begin{equation}
%\small
$s_{\text{new}} = max\{\lfloor \beta \times {s_{\text{freq}}}/ {s_{\text{rare}}}\rfloor, 1\} \times s_{\text{rare}}$,
%\end{equation}
where $s_{\text{freq}}$ and $s_{\text{rare}}$ are the numbers of samples in frequent and rare classes in a mini-batch, respectively, and $\lfloor \cdot \rfloor$ is the floor function. Finally, the feature maps of newly generated samples are concatenated with the original input feature maps along the batch dimension and forwarded to subsequent layers to calculate the loss and to optimize the whole framework.

\section{Experimental Evaluation}
%\footnote{The implementation details are given in the supplementary material.}
\label{experiment}
\paragraph{Datasets.} The experimental evaluation focuses on the Imabalanced CIFAR, the iNaturalist 2018, the  Places-LT, and the  ImageNet-LT datasets. Imbalanced CIFAR is based on the original CIFAR dataset, which is constructed by reducing the training samples per class, and the validation set is not changed. An imbalance ratio $\rho$ is defined as the ratio between sample sizes of the most frequent class and the least frequent class, i.e., $\rho = {N_{\text{max}}}/{N_{\text{min}}}$. We conducted experiments on the long-tailed imbalance \cite{cui2019class} and step imbalance~\cite{buda2018systematic} settings. The imbalance factors ($\rho$) that we used in our  experiments are 50 and 100. The iNaturalist species classification dataset \cite{van2018inaturalist} is a large-scale imbalanced dataset of 437,513 training images  classified into 8142 species in its 2018 version. The official training and validation set has a long-tailed distribution and  a  balanced distribution, respectively. 
Places-LT has 365 categories, with the maximum of 4980 images per class and the minimum of 5 images per class, while ImageNet-LT has 1000 categories, with the maximum of 1280 images per class and the minimum of 5 images per class. 
As for the evaluation on these two datasets, the classes are further categorized into three splits: many-shot (more than 100 samples), medium-shot (between 20 to 100), and few-shot (less than 20) in order to better examine performance variations across classes with different numbers of samples seen during training. We follow the experimental setting of these datasets in previous works \cite{cao2019learning, kang2019decoupling} for evaluation.

\vspace*{-1ex}
\paragraph{Implementation details.} The training details on the four datasets are summarized as follows:

\begin{itemize}[leftmargin=8pt]
\item {\bf Imbalanced CIFAR:} We followed the basic data augmentation method \cite{he2016deep} for training: 4 pixels are padded, and a $32\times32$ patch is randomly cropped from the image or its horizontal flip. The framework was trained with a batch size of 128 for 200 epochs. The learning rate was initially set to~0.1, and then it was decayed by~0.01 at the 160-th epoch and again at the 180-th epoch. The network was optimized by using stochastic gradient descend with a momentum of 0.9.

\item {\bf iNaturalist 2018:} We followed standard practice and performed data augmentation with random-size cropping \cite{szegedy2015going} to $224\times224$ from images or their horizontal flip. The network was trained from scratch for 90 epochs with a batch size of 256. The learning rate was set to 0.1 initially, and then it was decayed by 0.1 at the 50-th epoch, the 70-th epoch, and the 85-th epoch, respectively. Besides, for a fair comparison, we followed Kang \emph{et al.} \cite{kang2019decoupling} and also trained the model for the $2\times$ schedular (180 epochs). In our $2\times$ schedular experiment, the learning rate was decayed by 0.1 at the 100-th epoch, the 140-th epoch, and the 170-th epoch, respectively. During validation, images were center-cropped to $224\,{\times}\,224$ without further augmentation.

\item {\bf Places-LT:} We followed previous work \cite{liu2019large} to perform the data augmentation and to fine-tune ResNet-152, which is pre-trained on the full ImageNet-2012 dataset. The network was trained with a batch size of 256 for 30 epochs. The initial learning rate was set to 0.01, and it was decayed by 0.1 at every 10 epoch, and the training was stopped after 30 epochs. 

\item {\bf ImageNet-LT:}  We followed previous work \cite{kang2019decoupling} to use ResNeXt-50-32x4d, which was trained with a batch size of 256 for 100 epochs. The initial learning rate was set to 0.1, and it was decayed by 0.1 at the 60-th epoch, the 80-th epoch, and the 95-th epoch, respectively.

\end{itemize}

\begin{table}
\centering
\footnotesize
\resizebox{0.455\textwidth}{!}{
\begin{tabular}{@{}ccccc@{}}
 \toprule[1pt]
 \multirow{2}{*}{{\bf CIFAR-10}}    & \multicolumn{2}{c}{Long-Tailed} &\multicolumn{2}{c}{Step} \\
 \cline{2-5}
 & w/o RSG & w/ RSG  & w/o RSG & w/ RSG   \\
 \hline
 ERM & 25.19 &  {\bf 20.25}  & 28.88 & {\bf 26.07}   \\
 Focal Loss \cite{lin2017focal}  & 23.28 & {\bf 21.58} & 28.70 & {\bf 26.01}    \\
 M-DRW \cite{wang2018additive,cao2019learning} & 20.44 & {\bf 17.72} & 21.05 & {\bf 20.09}  \\
 LDAM-DRW \cite{cao2019learning} & 18.97 & {\bf 17.20} & 18.67 & {\bf 17.90}   \\
 \midrule[1pt]
    \multirow{2}{*}{{\bf CIFAR-100}}  & \multicolumn{2}{c}{Long-Tailed} &\multicolumn{2}{c}{Step} \\
    \cline{2-5}
   & w/o RSG & w/ RSG  & w/o RSG & w/ RSG   \\
 \hline
 ERM   & 56.15   &  {\bf 54.44 } & 59.32 & {\bf 56.82 }   \\
 Focal Loss \cite{lin2017focal} & 55.68  & {\bf 54.85 } & 58.50 & {\bf 55.93 }    \\
 M-DRW \cite{wang2018additive, cao2019learning} & 56.06  & {\bf 55.30 } & 56.26  & {\bf 54.60 } \\
 LDAM-DRW \cite{cao2019learning} &  53.38 & {\bf 51.50 } & 50.97  & {\bf 49.43 }   \\
  \bottomrule[1pt]
  \end{tabular}}
  \vspace{-0.2cm}
  \caption{Top-1 error rates of ResNet-32 with RSG for different loss functions on Imbalanced CIFAR for $\rho=50$.}
   \vspace{-0.5ex}
 \label{tab:loss_compare}
 \end{table}
 
\vspace*{-1ex}
\paragraph{Ablation studies.}
We performed ablation studies on  Imbalanced CIFAR with $\rho=50$. The mean error rates that are taken from three independent runs are reported. We comprehensively searched the hyperparameters of RSG and explored which level of feature is the most suitable for RSG to generate new samples by conducting experiments on ResNet-32 \cite{he2016deep} with LDAM-DRW \cite{cao2019learning},  where ``DRW'' denotes a deferred re-weighting training strategy proposed by Cao~\emph{et al.}  \cite{cao2019learning}. Based on our exploration, in the following experiments, we set the number of centers to 15, the frequent-class ratio to 0.2 and 0.5
for long-tailed and step imbalanced distributions, the transfer strength to 1.0 and 0.01 for long-tailed and step imbalanced distributions, and  $\lambda_1$ and $\lambda_2$ to 0.1 and 0.01, respectively. The search process can be found in the supplementary material. Note that RSG was initially used before the second-to-last down-sampling layer.

\begin{table}
 \centering
\resizebox{0.455\textwidth}{!}{
 % \begin{tabular}{cp{8mm}p{8mm}p{8mm}p{8mm}}
 \begin{tabular}{@{}ccccc@{}}
 \toprule[1.0pt]
 \multirow{2}{*}{{\bf CIFAR-10}}    & \multicolumn{2}{c}{Long-Tailed} &\multicolumn{2}{c}{Step} \\
 \cline{2-5}
 & w/o RSG & w/ RSG  & w/o RSG & w/ RSG   \\
 \hline
 ResNet-32  & 18.97 & {\bf 17.20} & 18.67 & {\bf 17.90}   \\
 ResNet-56  & 18.01 & {\bf 16.83} & 18.52 & {\bf 17.20}    \\
 ResNet-110  & 17.70 & {\bf 16.61} & 17.96 & {\bf 16.73}  \\
 DenseNet-40  & 17.46 & {\bf 16.21} & 17.40  & {\bf 16.12}   \\
 ResNeXt-29, 8$\times$64d & 16.10 & {\bf 15.26} & 16.82 & {\bf 15.99}  \\
 \midrule[1.0pt]
    \multirow{2}{*}{{\bf CIFAR-100}}  & \multicolumn{2}{c}{Long-Tailed} &\multicolumn{2}{c}{Step} \\
    \cline{2-5}
   & w/o RSG & w/ RSG  & w/o RSG & w/ RSG   \\
 \hline
 ResNet-32  & 53.38 &  {\bf 51.50} & 50.97 & {\bf 49.43}   \\
 ResNet-56  & 51.63 & {\bf 50.60} & 49.22& {\bf 48.53}    \\
 ResNet-110  & 50.64 & {\bf 49.83} & 48.65 & {\bf 47.90}  \\
 DenseNet-40  & 49.51 & {\bf 48.75} & 48.30 & {\bf 47.13}   \\
 ResNeXt-29, 8$\times$64d & 49.62 & {\bf 48.70} & 50.68 & {\bf 47.16}   \\
 \bottomrule[1.0pt]
 \end{tabular}}
  \vspace{-0.2cm}
  \caption{Top-1 error rates of different network architectures combined with LDAM-DRW \cite{cao2019learning} on Imbalanced CIFAR for $\rho=50$.}
 \label{tab:arch_compare}
  \vspace{-0.5ex}
\end{table} 

Firstly, we fixed the network architecture to ResNet-32 \cite{he2016deep} and tested RSG relative to different $L_{\text{cls}}$.
By Table~\ref{tab:loss_compare}, the deep model equipped with RSG consistently performs better than the one without RSG when combined with different loss functions. RSG significantly improves the performance when the model is combined with standard softmax with cross-entropy loss (denoted ERM, i.e., empirical risk minimization). This is reasonable, as standard softmax does not have any mechanism against imbalanced datasets. As for focal loss, AM-softmax, and LDAM,  although they are well-designed to tackle imbalanced datasets, RSG can still further improve the performance.

Secondly, we set $L_{\text{cls}}$ to LDAM with DRW \cite{cao2019learning} (i.e., LDAM-DRW) and evaluated (mainly five) different network architectures combined with RSG on Imbalanced CIFAR, namely, ResNet-32, ResNet-56, ResNet-110, DenseNet-40, and ResNeXt-29 (8$\times$64d).
Note that the used networks %that we used in this experiment 
were built according to the experiments on CIFAR in their original papers \cite{he2016deep, xie2017aggregated, huang2017densely}. 
As Table~\ref{tab:arch_compare} shows, when RSG is integrated into the networks, all the models are consistently improved.

Thirdly, we did a comprehensive ablation study on MV loss and the vector transformation module, and we obtain the following conclusions based on Table~\ref{tab:add_compare}: (1) Every subterm of MV loss is important and useful, since once we remove any subterm of it, an increase can be observed with regard to the error rate. (2) Adding the {feature displacement} to the centers of rare classes leads to an increase in terms of the error rate. This fact verifies what we have mentioned in Section~\ref{model}, i.e., adding the {feature displacement} to real rare-class samples is a better choice than adding it to the centers of rare classes. (3) Using the vector transformation module with MV loss performs better than directly adding the {feature displacement} to the samples in rare classes, which thus verifies their effectiveness.

Moreover, RSG is compared with previous sample generation methods \cite{schwartz2018delta, wang2018low, yin2019feature}. As Table~\ref{tab:compare_prev} shows, RSG has outperformed previous methods with different margins, showing that RSG can solve the drawbacks in previous generation methods and improve the performance.

Finally, we leveraged RSG before different pooling layers of ResNet-32 to explore which level of feature is the most suitable for generating new samples. As Table~\ref{tab:diff_layers} shows, RSG achieves the best result when it was used before the second-to-last down-sampling layer. Therefore, in the remaining experiments,  RSG was still used before the second-to-last down-sampling layer.

\begin{table}
\centering
\resizebox{0.475\textwidth}{!}{
\begin{tabular}{@{}ccccc@{}}
 \toprule[1pt]
 \multirow{2}{*}{}    & \multicolumn{2}{c}{Long-Tailed} &\multicolumn{2}{c}{Step} \\
 \cline{2-5}
 & CIFAR-10 & CIFAR-100 & CIFAR-10 & CIFAR-100   \\
 \hline
 MV Loss w/o 1st Term & 18.03 &  52.15 &  18.58 & 50.34  \\
 MV Loss w/o 2nd Term & 18.07 &  52.33 & 18.36  & 50.19 \\
 MV Loss w/o 3rd Term & 17.67 &  52.12 & 18.23  & 49.84 \\
 Adding to Rare-class Centers & 18.91 &  52.79 & 18.47  & 50.67 \\
 Direct Addition & 18.87 &  52.48 & 18.33  & 49.99  \\
 Vector Transformation Module  &  {\bf 17.20} & {\bf 51.50} &  {\bf 17.90}  &  {\bf 49.43}  \\
  \bottomrule[1pt]
  \end{tabular}}
 \vspace{-0.2cm}
 \caption{Ablation study on MV loss and the vector transformation module.
 Top-1 error rates of ResNet-32 combined with RSG and LDAM-DRW \cite{cao2019learning} on Imbalanced CIFAR for $\rho=50$ are reported.} %\vspace{-0.5ex}
 \label{tab:add_compare}
 \end{table}

\begin{table}
\centering
\resizebox{0.475\textwidth}{!}{
\begin{tabular}{@{}ccccc@{}}
 \toprule[1.0pt]
 \multirow{2}{*}{}    & \multicolumn{2}{c}{Long-Tailed} &\multicolumn{2}{c}{Step} \\
 \cline{2-5}
 & CIFAR-10  & CIFAR-100  & CIFAR-10  & CIFAR-100   \\
 \hline
 $\Delta$-Encoder \cite{schwartz2018delta} & 23.76 & 54.91 & 27.70 & 57.85  \\ 
 Imaginary \cite{wang2018low} & 23.99 & 55.08 & 28.23 & 58.46 \\
 FTL  \cite{yin2019feature} & 23.56 & 55.24 & 27.83 & 58.03 \\
 ERM-RSG (ours)  & {\bf 20.25} & {\bf 54.44} & {\bf 26.07} & {\bf 56.82} \\
 \bottomrule[1.0pt]
 \end{tabular}}
 \vspace{-0.2cm}
 \caption{Comparison with other sample generation methods on Imbalanced CIFAR ($\rho=50$). All of them are based on ResNet-32 combined with ERM for a fair comparison.}
 %\vspace{-0.5ex}
 \label{tab:compare_prev}
 \end{table}

\begin{table}[h]
\centering 
\resizebox{0.48\textwidth}{!}{
\begin{tabular}{ccccc}
 \toprule[1.0pt]
 \multirow{2}{*}{}    & \multicolumn{2}{c}{Long-Tailed} & \multicolumn{2}{c}{Step} \\
\cline{2-5}
 & CIFAR-10  & CIFAR-100  & CIFAR-10  & CIFAR-100   \\
 \hline
 1st down-sampling & 18.13 & 53.22 & 18.66 & 50.81   \\
 2nd down-sampling & {\bf17.20} & {\bf51.50} & {\bf17.90} & {\bf49.43}  \\
 3rd down-sampling (GAP)  & 17.68 &  52.14& 18.05 & 50.38  \\
 \bottomrule[1.0pt]
 \end{tabular}}
 \caption{Ablation study (top-1 error rates) with regard to the different layers, where RSG was used on Imbalanced CIFAR ($\rho=50$). RSG was used before the three down-sampling layers in ResNet-32. ResNet-32 combined with LDAM-DRW was used, and GAP denotes global average pooling.}
 \label{tab:diff_layers}
 \end{table}

 \vspace{-0.5cm}
\paragraph{Comparison with state of the art.}
For each of the following experiments, we report mean error rates or mean accuracies, which are taken from three independent runs. Table~\ref{tab:soa_cifar} shows the results on Imbalanced CIFAR with $\rho\,{\in}\,\{50$, $100\}$. We first compare our LDAM-DRW-RSG with LDAM-DRW, as this comparison directly shows the improvement brought by RSG. After combining LDAM-DRW with RSG, we obtain a remarkable improvement for both long-tailed  and step imbalanced 
distributions, which shows the power of RSG for handling imbalanced datasets. As a result, with the help of RSG, LDAM-DRW-RSG achieves superior results on Imbalanced CIFAR when compared with previous methods. 

\begin{table*}
\centering\resizebox{0.75\textwidth}{!}{
 \begin{tabular}{@{}c|c|c|c|c|c|c|c|c@{}}
 \hline
 Dataset  & \multicolumn{4}{c|}{Imbalanced CIFAR-10} &\multicolumn{4}{c}{Imbalanced CIFAR-100} \\
 \hline
 Imbalance Type    & \multicolumn{2}{c|}{Long-Tailed} &\multicolumn{2}{c|}{Step} & \multicolumn{2}{c|}{Long-Tailed} &\multicolumn{2}{c}{Step}\\
 \hline
 Imbalance Ratio ($\rho$) & 100 & 50  & 100 & 50 & 100 & 50 & 100 & 50  \\
 \hline
  ERM  & 29.64  & 25.19  & 36.70 & 28.88 & 61.68 & 56.15 & 61.43 & 59.32\\
  Focal loss \cite{lin2017focal} & 29.62 & 23.28 & 36.09 & 28.70 & 61.59 & 55.68 & 61.65 & 58.50\\
  CB Focal \cite{cui2019class} & 25.43  & 20.73  & 39.73 & 39.65 & 63.98 & 54.83 & 80.24 & 85.10 \\
  CB RW \cite{cui2019class} & 27.63  & 21.95  & 38.06 & 30.38 & 66.01 & 57.54 & 78.69 & 69.63 \\
  M-DRW \cite{cao2019learning} & 24.94  & 20.44  & 27.67 & 21.05 & 59.49 & 56.06  & 58.91 & 56.26\\
  BBN \cite{zhou2020BBN}& {\bf 20.18} & 17.82 & 22.34  & 18.33 & 57.44 & 52.98 & 54.14 & 50.49 \\
  LDAM-DRW \cite{cao2019learning} & 22.97 & 18.97 & 23.08  & 18.67 & 57.96 & 53.38 & 54.64 & 50.97 \\
  LDAM-DRW-SSP \cite{yang2020rethinking} & 22.17 & 17.87 & 22.95   & 18.38  & 56.57 & 52.89 &  54.28 &  50.47 \\
  LDAM-DRW-RSG (ours) & 20.45  & {\bf 17.20} &  {\bf 21.65}  & {\bf 17.90} & {\bf 55.45}  & {\bf 51.50} & {\bf 53.00} & {\bf 49.43}\\
 \hline
 \end{tabular}}\vspace{-1.2ex}
 \caption{Top-1 error rates of ResNet-32 on Imbalanced CIFAR.}
 \label{tab:soa_cifar}
  %\vspace{-0.3ex}
  \vspace{-2ex}
\end{table*}

\begin{table}
\centering
\resizebox{0.45\textwidth}{!}{
\begin{tabular}{@{}c|c|c@{}}
 \hline
Training Schedular & Method & Error Rate    \\
   \hline
\multirow{11}{*}{$1 \times$ schedular}  & ERM  & 42.86     \\
  & CB Focal Loss \cite{cui2019class} & 38.88      \\
  & ERM-DRW \cite{cao2019learning} & 36.27   \\
  & ERM-DRS \cite{cao2019learning} & 36.44 \\
  & BBN \cite{zhou2020BBN} & 33.71 \\
  & $\tau$-normalized \cite{kang2019decoupling} & 34.40   \\
  & LDAM-DRW \cite{cao2019learning} &  34.00   \\
  & LDAM-DRS \cite{cao2019learning} &  32.73   \\
  & LDAM-DRW-SSP \cite{yang2020rethinking} &  33.70   \\
  & LDAM-DRW-RSG (ours) &  33.22 \\
  & LDAM-DRS-RSG (ours) &  {\bf 32.10}  \\
 \hline
\multirow{5}{*}{$2 \times$ schedular}  & BBN \cite{zhou2020BBN}  & 30.38 \\
  &$\tau$-normalized  \cite{kang2019decoupling}& 30.70  \\
  & cRT \cite{kang2019decoupling} & 32.40 \\
  & LWS  \cite{kang2019decoupling} & 30.50  \\
  &LDAM-DRS-RSG (ours) & {\bf29.74} \\
 \hline
\end{tabular}}
\vspace{-0.2cm}
 \caption{Top-1 error rates of ResNet-50 on iNaturalist 2018.}
 \label{tab:soa_inaturalist}
\vspace{-0.5ex}
\end{table}

Table~\ref{tab:soa_inaturalist} shows the top-1 error rate of different methods using ResNet-50 \cite{he2016deep} as the backbone on iNaturalist 2018, and we followed Kang \emph{et al.} \cite{kang2019decoupling} to conduct experiments in two training settings, namely, the 1$\times$ schedular and the 2$\times$ schedular.
%\footnote{Details about training schedular settings can be found in the implementation details in the supplementary material.} 
In the 1$\times$ schedular experiment, we compare LDAM-DRW-RSG and LDAM-DRS-RSG with previous LDAM-DRW and LDAM-DRS, separately. Here, ``DRS'' denotes a  deferred class-balanced resampling strategy proposed by Cao \emph{et al.} \cite{cao2019learning}. Note that we cannot reproduce the result on iNaturalist 2018 reported in the original paper (32.0\%) \cite{cao2019learning} by using LDAM-DRW. So, we report our reproduced results of LDAM-DRW and LDAM-DRS \cite{cao2019learning} based on their publicly available code. The results in Table~\ref{tab:soa_inaturalist} show that we can obtain better results by leveraging the proposed generator, which directly demonstrates the effectiveness of RSG. Moreover, as for the 2$\times$ schedular setting, the top-1 error rate of  LDAM-DRS-RSG is further decreased. Thus, it can be seen that RSG helps the model achieve new state-of-the-art results in both training schedular settings, which demonstrates that RSG is capable of dealing with imbalanced datasets~effectively.

Table~\ref{tab:soa_places} shows the top-1 accuracy on Places-LT. The results show that the performance can be further improved when RSG is combined with LDAM-DRS, showing that RSG is useful. Moreover, when compared with the recent two popular methods, namely,  $\tau$-normalized \cite{kang2019decoupling} and BBN \cite{zhou2020BBN}, RSG can improve the performance of the model on medium-shot and few-shot classes with less accuracy loss on many-shot classes, resulting in a higher overall accuracy and a new state-of-the-art result. 

Table~\ref{tab:soa_imagenet} shows the top-1 accuracy on ImageNet-LT. When compared with LDAM-DRW, LDAM-DRW-RSG can achieve a higher accuracy, verifying that RSG is able to alleviate the problem caused by imbalanced datasets. RSG can enhance the model and greatly improve its generality on medium-shot and few-shot classes. In addition, by equipping RSG, we can also obtain a new state-of-the-art result on ImageNet-LT.

Since all hyperparameters of RSG were fixed after the hyperparameter searching process, we can conclude that the hyperparameters and RSG are quite robust to new datasets (i.e., Places-LT,  ImageNet-LT, and iNaturalist 2018). If hyperparameters are further tuned on the new datasets, even better results might be obtained.

\begin{table}
\centering
\resizebox{0.45\textwidth}{!}{
\begin{tabular}{@{}ccccccc@{}}
     \hline
 Method & Many & Medium  & Few & All\\
 \hline
 Lifted Loss \cite{oh2016deep} & 41.1 &  35.4 & 24.0 & 35.2\\
 Focal Loss \cite{lin2017focal} & 41.1 & 34.8 & 22.4 & 34.6\\
 Range Loss \cite{zhang2017range} & 41.1  & 35.4 & 23.2 & 35.1\\
 FSLwF \cite{gidaris2018dynamic} & 43.9 & 29.9 & 29.5 & 34.9 \\
 BBN \cite{zhou2020BBN} & 42.5 & 40.3 & 30.6 & 38.7\\
 OLTR \cite{liu2019large} &  {\bf 44.7} & 37.0 & 25.3 & 35.9 \\
 $\tau$-normalized \cite{kang2019decoupling} & 37.8& 40.7 & 31.8 & 37.9\\
 LDAM-DRS \cite{cao2019learning} & 43.3 & 38.3 & 30.7 & 38.6 \\
 LDAM-DRS-RSG (ours) & 41.9& {\bf 41.4} & {\bf 32.0} & {\bf 39.3}\\
 \hline
\end{tabular}}
\vspace{-0.2cm}
  \caption{Top-1 accuracy of ResNet-152 on Places-LT.}
  \label{tab:soa_places}
  %\vspace{-0.5ex}
\end{table}

\begin{table}
\centering
\resizebox{0.45\textwidth}{!}{
\begin{tabular}{@{}ccccccc@{}}
     \hline
 Method & Many & Medium  & Few & All\\
 \hline
 Focal Loss \cite{lin2017focal} & 63.3 & 37.4 & 7.7 & 43.2 \\
 OLTR \cite{liu2019large} &52.1 & 39.7 & 20.3 & 41.2 \\
 Joint \cite{kang2019decoupling} & {\bf 65.9} & 37.5 & 7.7 & 44.4 \\
 NCM \cite{kang2019decoupling} & 56.6 &  45.3  & 28.1 &  47.3 \\
 cRT \cite{kang2019decoupling} & 61.8 &  46.2 &  27.4  &  49.6 \\
 $\tau$-normalized \cite{kang2019decoupling} & 59.1 & 46.9 & 30.7 & 49.4 \\
 LWS \cite{kang2019decoupling} & 60.2 & 47.2 & 30.3 & 49.9 \\
 LDAM-DRS \cite{cao2019learning} & 63.7 & 47.6 & 30.0 & 51.4 \\
 LDAM-DRS-RSG (ours) & 63.2 & {\bf 48.2} & {\bf 32.3} & {\bf 51.8 }\\
 \hline
\end{tabular}}
\vspace{-0.2cm}
  \caption{Top-1 accuracy of ResNeXt-50 on ImageNet-LT.}
  \label{tab:soa_imagenet}
  \vspace{-1.2ex}
\end{table}

\section{Summary and Outlook}

\label{conclusion}
%In this work, w
We have introduced a rare-class sample generator (RSG), which is a general building block to mitigate the issue of training on imbalanced datasets. RSG is simple yet effective,  since it is an architecture-agnostic and loss-agnostic plug-in module, and it does not bring any additional burdens to the backbone network during the inference phase. In extensive experiments, we have verified the effectiveness of RSG, which has achieved excellent results on four public benchmarks. Since RSG is flexible and orthogonal to most previous methods, future research can focus on improving the RSG module directly by designing more elegant ways to generate higher-quality rare-class samples.

\vspace{-1ex}
\paragraph{\small Acknowledgments.} \small This work was supported by the National Natural Science Foundation of China under the grant 61906063, by the Natural Science Foundation of Tianjin City, China, under the grant 19JCQNJC00400, and by the ``100 Talents Plan'' of Hebei Province, China, under the grant E2019050017.
This work was also supported by the Alan Turing Institute under the EPSRC grant EP/N510129/1 and
by the AXA Research Fund. We also acknowledge the use of the Tier 2 facility
JADE (EP/P020275/1) and GPU computing support by Scan Computers International Ltd.

{\small
\bibliographystyle{ieee_fullname}
\bibliography{egbib}
}

\end{document}